\documentclass{article}

\usepackage{PRIMEarxiv}

\usepackage[utf8]{inputenc} % allow utf-8 input
\usepackage[T1]{fontenc}    % use 8-bit T1 fonts
\usepackage{hyperref}       % hyperlinks
\usepackage{url}            % simple URL typesetting
\usepackage{booktabs}       % professional-quality tables
\usepackage{amsfonts}       % blackboard math symbols
\usepackage{nicefrac}       % compact symbols for 1/2, etc.
\usepackage{microtype}      % microtypography

\usepackage{lipsum}
\usepackage{xcolor}
\usepackage{fancyhdr}       % header
\usepackage{graphicx}       % graphics
\graphicspath{{media/}}  
\usepackage{natbib}
\usepackage{amsmath}
\usepackage{soul}
\usepackage{amssymb}
\usepackage{booktabs}% organize your images and other figures under media/ folder

\title{On Global Applicability and Location Transferability of Generative Deep Learning Models for Precipitation Downscaling
}

\author{
Paula Harder \\
ECMWF \& Mila Quebec AI Institute\\
\And
  Christian Lessig, Matthew Chantry \\
  ECMWF \\
   \And
  Francis Pelletier, David Rolnick \\
  Mila Quebec AI Insitute \\
    }

\begin{document}
\maketitle

\begin{abstract}
Deep learning offers promising capabilities for the statistical downscaling of climate and weather forecasts, with generative approaches showing particular success in capturing fine-scale precipitation patterns. However, most existing models are region-specific, and their ability to generalize to unseen geographic areas remains largely unexplored. In this study, we evaluate the generalization performance of generative downscaling models across diverse regions. Using a global framework, we employ ERA5 reanalysis data as predictors and IMERG precipitation estimates at $0.1^\circ$ resolution as targets. A hierarchical location-based data split enables a systematic assessment of model performance across 15 regions around the world.
\end{abstract}


\section{Introduction}

Accurate modeling of weather and climate is critical for understanding global climate dynamics and supporting adaptation strategies in sectors such as agriculture, infrastructure, and renewable energy. However, global climate and weather models typically operate at coarse spatial resolutions of 10–80 km, which limits their ability to capture fine-scale processes such as extreme rainfall, topographic effects, and coastal dynamics. High-resolution data are essential not only for assessing the impacts of extreme events but also for optimizing renewable energy production and informing adaptation measures.

Downscaling addresses this limitation by deriving high-resolution information from coarse-resolution model outputs. This is particularly important for precipitation, where rainfall patterns can vary substantially at scales below those resolved by global models. Conceptually, downscaling resembles super-resolution in computer vision, but with important differences: it must correct biases in the input data and often leverages multiple large-scale predictors to estimate fine-scale targets, a process known as heterogeneous downscaling.

A widely used strategy is stochastic downscaling, which generates ensembles of high-resolution realizations consistent with large-scale dynamics while capturing small-scale uncertainties such as terrain influences and coastal effects. Such probabilistic approaches are especially suitable, as the mapping between coarse and fine scales is inherently uncertain and cannot be fully resolved with deterministic models given computational constraints.

The effectiveness of deep learning methods for downscaling is strongly dependent on the availability and distribution of high-quality training data. Although large volumes of climate and weather data exist, observational coverage is geographically imbalanced: often data are scarce in the Global South, for example, ground-based radar precipitation measurements. This imbalance poses challenges for generalization, as precipitation processes vary widely across regions; for example, convection-driven rainfall in equatorial areas differs fundamentally from precipitation regimes in Europe or North America.

In this paper, we investigate the generalization of a generative downscaling model to unseen locations. We focus on precipitation downscaling using an architecture based on \citet{Harris2022AGD}. For global evaluation, we construct a downscaling task from ERA5 predictors at $0.25^\circ$ resolution to precipitation from the satellite-based IMERG product. The globe is divided into nine subregions, from which we define hierarchical datasets yielding 15 training areas. We analyze if and how model performance degrades when applied outside the training regions, with particular emphasis on the role of changing orography.

\section{Related Work}

Deep learning is increasingly used for climate and weather downscaling, particularly in probabilistic precipitation downscaling \citep{EnhancingRegionalClimateDownscalingThroughAdvancesinMachineLearning}.
For a recent overview we refer to \citet{EnhancingRegionalClimateDownscalingThroughAdvancesinMachineLearning}. Super-resolution CNNs, especially U-Nets \citep{Sha2020DeepLearningBasedGD, https://doi.org/10.1002/met.1961} and ResNets \citep{10.1145/3394486.3403366, 8588749}, are among the most common architectures for downscaling. While some standard computer vision models can be directly applied, adaptation is required to address certain Earth system challenges. Recent advancements for specific data types include explainable and interpretable methods \citep{https://doi.org/10.1029/2023MS003641, RAMPAL2022100525}, physics-constrained neural networks \citep{my_jmlr, geiss_hardin, gonzalez2023multi}, and arbitrary-resolution downscaling using Fourier neural operators \citep{yang2023fourier}. 

Precipitation downscaling presents unique challenges, particularly in capturing stochastic high-frequency variations. Generative models have been most successful in this regard, with conditional generative adversarial networks (cGANs) \citep{Goodfellow2014GenerativeAN} being a popular choice \citep{leinonen2020stochastic, price2022increasingaccuracyresolutionprecipitation, Harris2022AGD}, especially stabilized variants like Wasserstein GANs \citep{Harris2022AGD, cooper2023analysiscgangenerativedeep}. More recently, diffusion-based models \citep{SohlDickstein2015DeepUL} have demonstrated strong performance in this domain \citep{Mardani2023GenerativeRD, Wan2023DebiasCS, addison2024machinelearningemulationprecipitation, ling_diffusion_2024}, leveraging their ability to model complex distributions. %In this paper, we consider state-of-the-art models for probabilistic precipitation downscaling -- GANs and diffusion-based models -- as baseline architectures.

Generalization across geographies is an active research area in applied ML, particularly in remote sensing and biodiversity modeling. In agricultural classification and segmentation, approaches such as task-informed meta-learning \cite{tseng2022timltaskinformedmetalearningagriculture}, versions of model-agnostic meta-learning \cite{rußwurm2020metalearning}, and multi-source unsupervised domain adaptation \cite{Wang31122022} have shown promise in adapting to new regions with minimal data. In biodiversity monitoring, \citet{NEURIPS2023_ef7653bb} integrate remote sensing and citizen science data to improve generalization in data-sparse regions like Kenya, while \citet{SINR_icml23} leverage spatial implicit neural representations for scalable global species range estimation using noisy, sparse data.
 However, in Earth system modeling geographical generalization remains underexplored. Initial studies have investigated the regional generalizability of downscaling models \cite{10440025}. For instance, some works examine generalization between different areas on the US West Coast \cite{Sha2020DeepLearningBasedGD, DeepLearningBasedGriddedDownscalingofSurfaceMeteorologicalVariablesinComplexTerrainPartIIDailyPrecipitation}. Others analyze performance across regions in the UK and the United States \cite{cooper2023analysiscgangenerativedeep} or evaluate generalization from the DACH region (Germany, Austria, and Switzerland) to North America \cite{prasad2024evaluatingtransferabilitypotentialdeep}. While these efforts provide valuable insights, their geographic scope remains limited. Here, we expand on this by providing the first worldwide evaluation for downscaling transferability.

\section{Data}

Our dataset combines ERA5 reanalysis data, IMERG satellite precipitation estimates, and high-resolution static geographic fields (orography and land–sea mask). ERA5 serves as the low-resolution (LR) input, while IMERG provides the high-resolution (HR) target for downscaling. The dataset covers nearly the entire globe between $60^\circ \text{N}$ and $60^\circ \text{S}$ — the region of complete IMERG coverage.

\subsection{Input: ERA5}

ERA5 \citep{era5} is the fifth-generation atmospheric reanalysis produced by the European Centre for Medium-Range Weather Forecasts (ECMWF). Reanalysis datasets combine historical observations with numerical weather prediction models to generate consistent, gridded estimates of the past state of the atmosphere. ERA5 provides hourly global fields at a spatial resolution of $0.25^\circ \times 0.25^\circ$ (approximately $25~\mathrm{km}$ per grid cell at mid-latitudes) on a regular latitude–longitude grid, spanning from 1950 to the present. To align with IMERG availability, we use data from 2001 onward.

Following \citet{Harris2022AGD}, originally informed by domain knowledge and the ecPoint model \citep{Hewson2020ALP}, we select nine predictor variables known to influence precipitation processes. These include precipitation types, atmospheric moisture content, and dynamic fields:

\begin{enumerate}
\item Total precipitation (tp)
\item Convective precipitation (cp)
\item Convective available potential energy (cape)
\item Total water content (twc)
\item Total liquid water content (tlwc)
\item Surface pressure (sp)
\item Top-of-atmosphere incident solar radiation (tisr)
\item Eastward wind component at 700 hPa (u)
\item Northward wind component at 700 hPa (v)
\end{enumerate}

\subsection{Input: Geographic Features}

To provide additional static context, we include two high-resolution ($0.1^\circ$) geographic fields:
(i) a land–sea mask indicating the land fraction within each grid cell, and
(ii) an orography map, represented as geopotential height at the surface.
These static variables are known to strongly influence local precipitation patterns, especially in regions with complex topography or coastal influences.

\subsection{Target: IMERG}

The Integrated Multi-satellite Retrievals for GPM (IMERG) product \citep{huffman_imerg_2014} is part of NASA’s Global Precipitation Measurement (GPM) mission. IMERG combines observations from the GPM satellite constellation with additional inputs such as gauge data to produce globally gridded precipitation estimates. It provides near-global coverage between $60^\circ \text{N}$ and $60^\circ \text{S}$ at $0.1^\circ$ spatial resolution (about $10~\mathrm{km}$ per pixel) and hourly temporal resolution. We use the IMERG V07 Final Run product \citep{huffman_imerg_v07} as the high-resolution target field.

\subsection{Preprocessing}

Minimal preprocessing is applied to the raw data, which are downloaded directly from publicly available sources. IMERG’s native half-hourly precipitation fields are averaged to hourly resolution to match ERA5. All data are then converted from NetCDF to NumPy array format to improve I/O efficiency during training. Alternative storage formats such as Zarr can be used if metadata retention is required.

\begin{figure*}[htb]
\centering
\includegraphics[width=\textwidth]{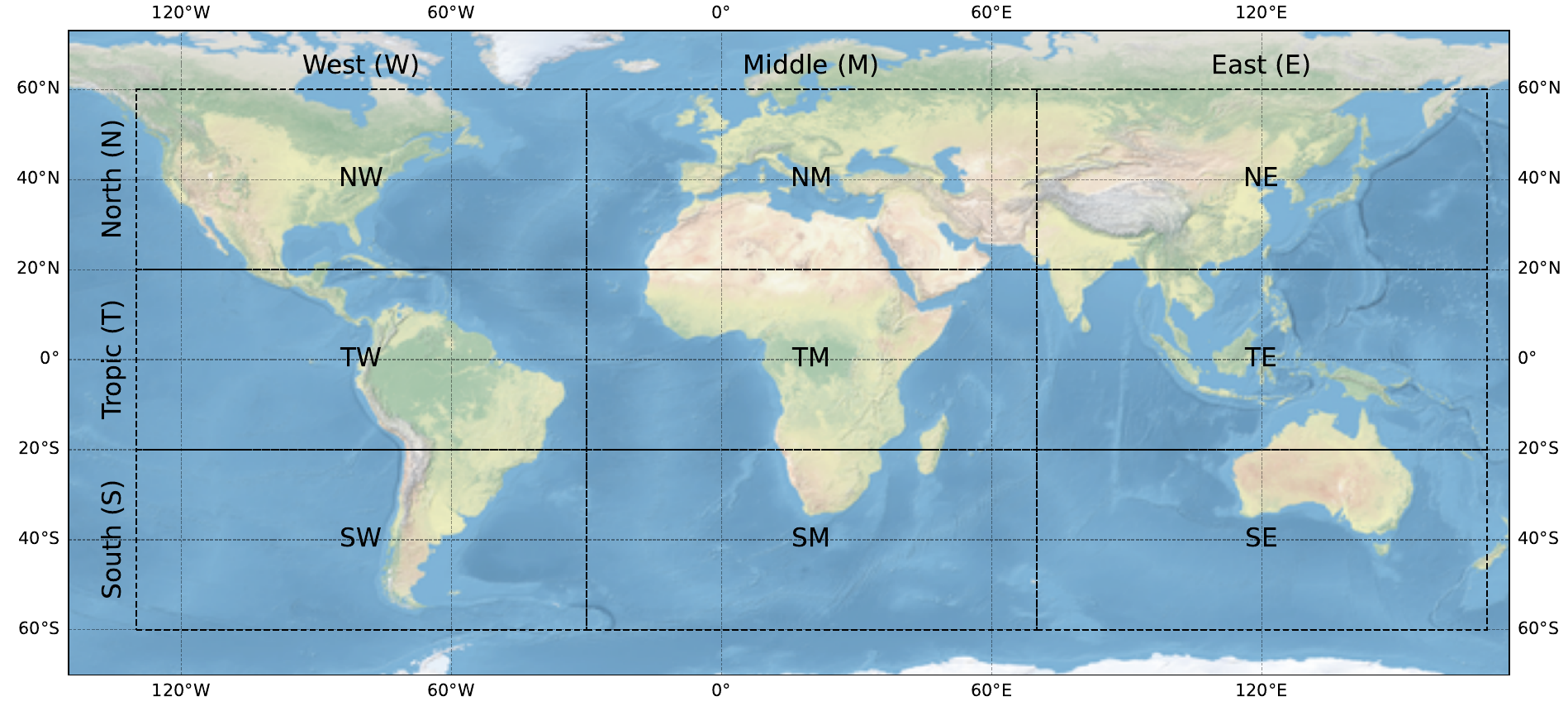}
\caption{Spatial domain and regional partitioning of the dataset. The covered area extends from $60^\circ$N to $60^\circ$S and $130^\circ$W to $170^\circ$E. We first divide the domain into nine rectangular subregions, which are then combined into larger regions (N, T, S, W, M, E) to create 15 training areas in total.}
\label{map}
\end{figure*}

Finally, the data are spatially partitioned into nine latitude–longitude subregions, as shown in Figure~\ref{map}. These subregions are further grouped into larger composite regions: north (N), tropics (T), south (S), west (W), middle (M), east (E), resulting in a total of 15 distinct training domains used in our experiments (see Section~\ref{exp} for details).

\section{Methodology}

For this work, we adopt the generative adversarial network (GAN) architecture proposed by \citet{Harris2022AGD} and adapt it to our precipitation downscaling task with minor modifications and targeted hyperparameter tuning.

\subsection{GAN Overview}

Generative Adversarial Networks (GANs) are a class of deep generative models that learn to synthesize data samples resembling a target distribution through a two-player minimax game. A generator network attempts to produce realistic high-resolution samples from low-resolution inputs, while a discriminator learns to distinguish generated outputs from real observations. 
To improve training stability and address common issues such as mode collapse, we use the Wasserstein GAN (WGAN) formulation \citep{arjovsky2017wasserstein}. WGAN replaces the standard adversarial objective with the Wasserstein distance.

\subsection{Model Architecture}

We employ the WGAN setup from \citet{Harris2022AGD} with minor adaptations. The training objective consists of three terms:
(i) A supervised content loss based on the mean squared error (MSE) between the mean of the generated ensemble and the target field.
(ii) An adversarial loss encouraging the generator to produce realistic fine-scale structures.
(iii) A gradient penalty term applied to the discriminator to ensure Lipschitz continuity.

The generator follows a ResNet-style architecture and takes as input the selected ERA5 variables and static geographic features. It outputs high-resolution precipitation fields at $0.1^\circ$ resolution. The discriminator (critic) receives three inputs — the generated sample, the low-resolution input, and the static fields — and is composed of residual blocks, average pooling layers, and a final dense classification head. Further architectural details can be found in \citet{Harris2022AGD}.

\paragraph{Hyperparameters}

Hyperparameters were selected through a combination of Bayesian optimization using the Weights \& Biases framework \citep{wandb} and manual fine-tuning. The search was conducted on a reduced-resolution setup (downsampling input and target resolution by a factor of four) to accelerate experimentation.
A key finding was that setting the generator learning rate higher than that of the discriminator improved convergence, contrary to the original configuration in \citet{Harris2022AGD}. Final hyperparameters are as follows: generator learning rate = 0.005, discriminator learning rate = 0.0005, content loss weight = 300, gradient penalty weight = 10, and ensemble size for the content loss = 8. Both networks use the Adam optimizer, with 128 filters in the generator and 256 filters in the discriminator.

\subsection{Baselines}

The superior performance of the WGAN relative to conventional baselines was demonstrated in \citet{Harris2022AGD}. In this study, we include a simple baseline — bilinear interpolation of total ERA5 precipitation — to quantify the relative performance of the downscaling model across regions.

\subsection{Training}

Each model is trained for five epochs. Given the dataset size, this corresponds to approximately 3–5 days of training on two NVIDIA A100 GPUs using distributed training via the Hugging Face Accelerate framework \citep{accelerate}.
Before training, both ERA5 precipitation variables (tp, cp) and IMERG precipitation targets are log-transformed using $\log(x + 10^{-5})$. All variables, including the transformed precipitation fields, are then normalized and standardized using the global mean and standard deviation computed per variable, except for the land–sea mask, which remains unchanged.

\subsection{Experiments} \label{exp}

As shown in Figure~\ref{map}, we divide the globe (from $60^\circ$N–$60^\circ$S and $130^\circ$W–$170^\circ$E) into nine equally sized latitude–longitude subregions. These are further combined into larger regions — North (N), Tropics (T), South (S), West (W), Middle (M), and East (E) — resulting in a total of 15 training domains. For each of these 15 regions, we train a separate GAN, which is subsequently evaluated on all nine subregions to assess generalization performance.

We additionally split the data temporally. Training is performed on the years 2001–2018. The years 2019 and 2020 are reserved for validation, hyperparameter tuning, and checkpoint selection. The final evaluation is conducted on 2021–2022 data, from which we compute Continuous Ranked Probability Scores (CRPS).

\subsection{Evaluation}
%evaluate 
Various metrics were investigated in \cite{Harris2022AGD}, here we focus on one major metric, the continuous ranked probability score (CRPS). 
The CRPS is a metric used to evaluate the accuracy of probabilistic forecasts. For a given forecast probability distribution $F$ and the observed outcome ${y}$, the CRPS is calculated as follows:
    \begin{equation}\label{eq:crps_definition}
 \text{CRPS}(F, {y}) = \int_{-\infty}^{\infty} [F(z) - \mathbf{1}(z \geq {y})]^2 \text{dz}.
\end{equation}
Here, $F(z)$ is the cumulative distribution function of the forecast distribution at point $z$ and
$\bf{1}(\cdot)$ the indicator function. For a deterministic forecast, the CRPS reduces to a mean-average error (MAE).

\section{Results and Discussion}

\paragraph{Total CRPS Scores}

\begin{figure*}[htb]
\centering
\includegraphics[width=\textwidth]{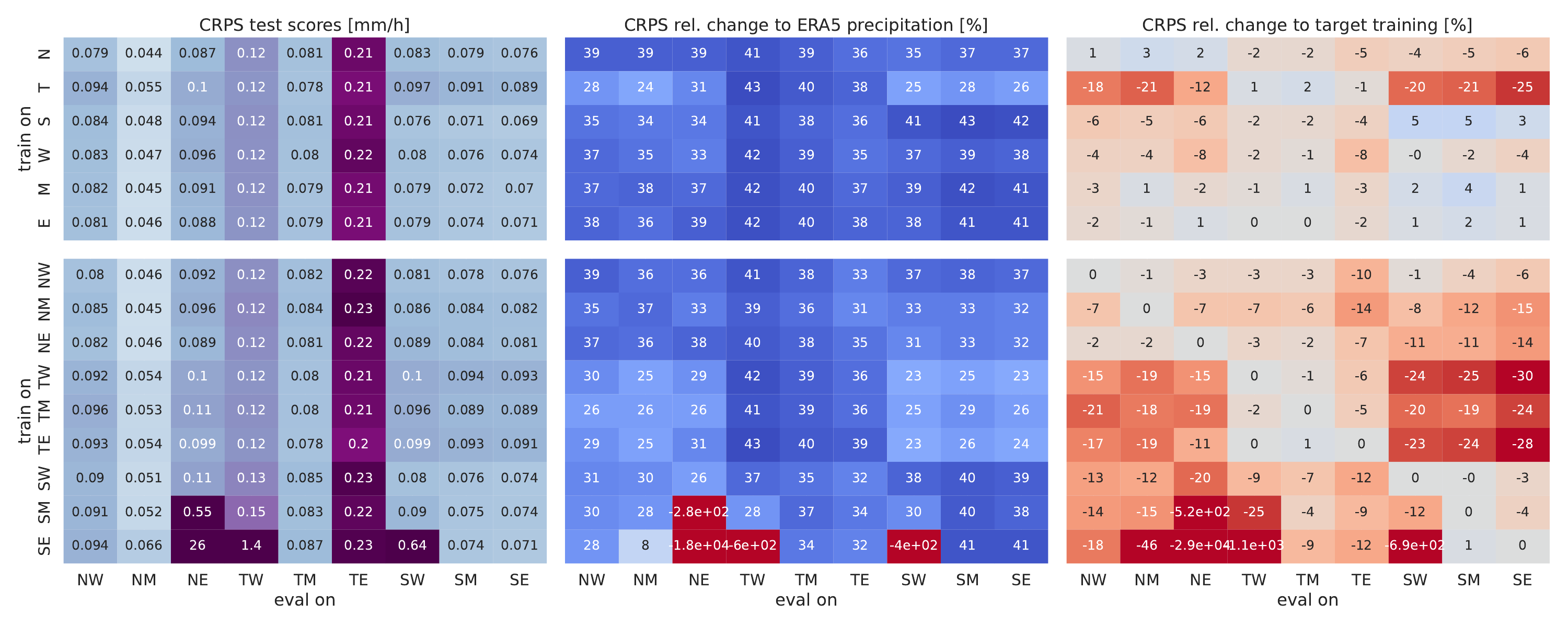}
\caption{CRPS scores for all training–evaluation region combinations. Rows correspond to training regions, columns to evaluation regions. Lower scores indicate better performance. The left column shows the overall CRPS performance.}
\label{tables}
\end{figure*}

\begin{figure*}[htb]
\begin{center}
\includegraphics[width=\textwidth]{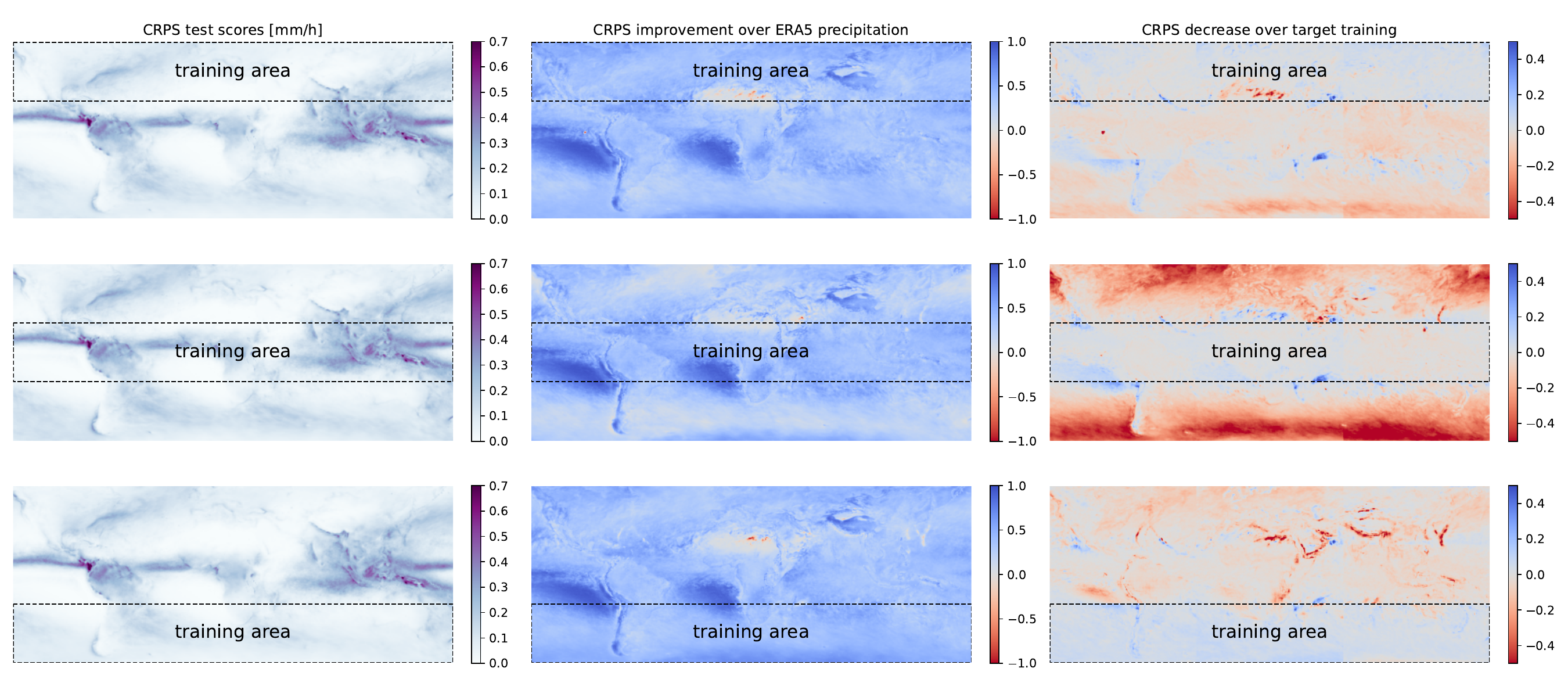}
\caption{CRPS score maps for three representative training regions: North (N), Tropics (T), and South (S). Left: raw CRPS. Middle: relative improvement over ERA5. Right: relative performance drop compared to direct training on the target region.}
\label{nts_maps}
\end{center}
\end{figure*}

\begin{figure*}[htb]
\begin{center}
\includegraphics[width=\textwidth]{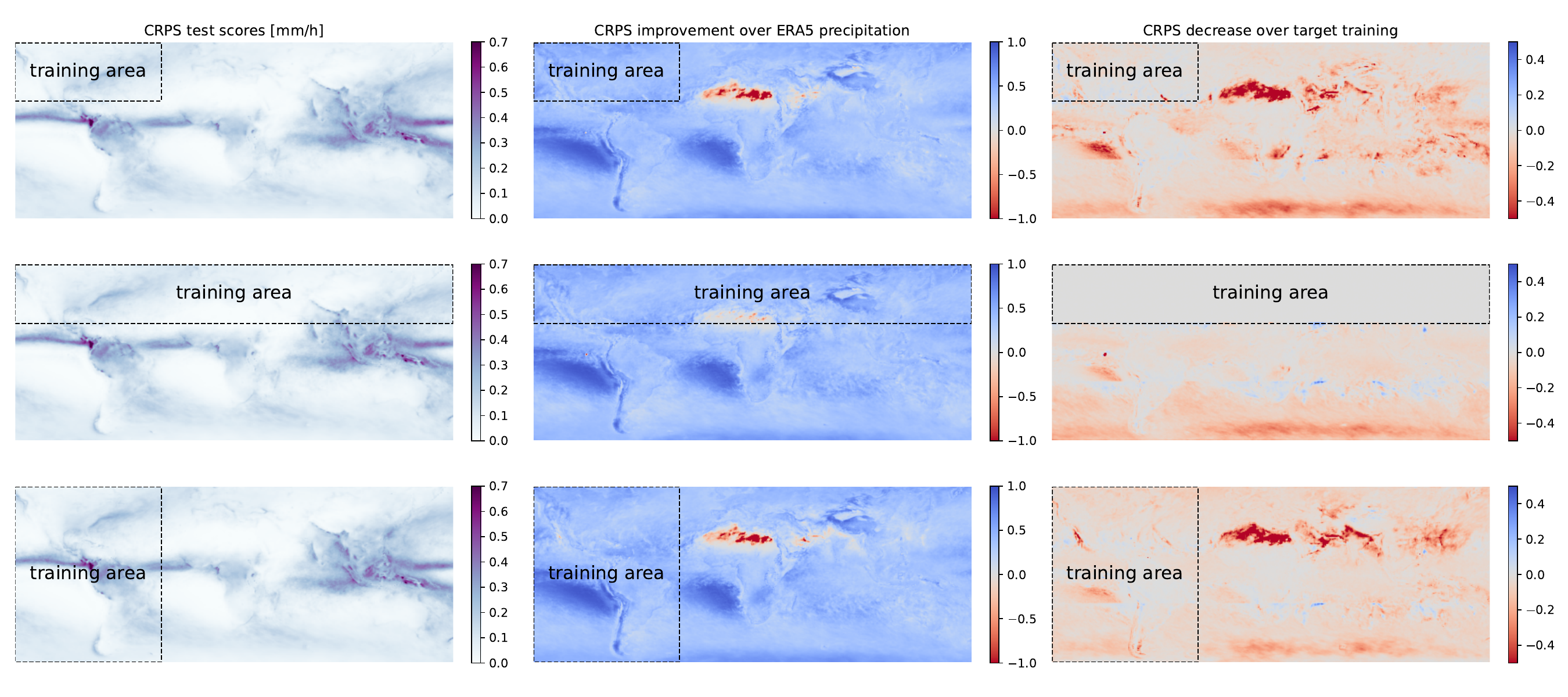}
\caption{As in Figure~\ref{nts_maps}, but for a hierarchical training setup: starting with Northwest (NW) and progressively enlarging the training domain. Enlarging the domain generally improves transfer performance, especially when the target region is included.}
\label{hierach_maps}
\end{center}
\end{figure*}

The overall CRPS performance across training–evaluation region pairs is shown in Figure~\ref{tables}. We observe substantial variability between evaluation regions, whereas differences between training locations evaluated at the same target location are comparatively minor. This is further illustrated by the leftmost column of Figures~\ref{nts_maps} and \ref{hierach_maps}, where the rows (corresponding to different training regions) show very similar spatial patterns.

The most challenging evaluation region are consistently in the tropics, which exhibits the highest CRPS scores — likely due to these region’s large precipitation amounts and strong convective activity. Some notable outliers highlight specific transfer difficulties: for example, models trained in the southeastern (Australia) region perform poorly in NE, TW, and SW — areas with complex topography. Similarly, the model trained on the South African region (SE) struggles in NE, which includes the Tibetan Plateau. These cases point to limitations in transferring models from low-relief to high-relief regions, discussed further below.

\paragraph{CRPS Improvement Over ERA5 Interpolation}

Compared to bilinear interpolation of ERA5 precipitation, the GAN achieves substantial performance gains in nearly all cases, consistent with findings from \citet{Harris2022AGD}. Except for five outliers, CRPS scores improve by 23–42\% relative to the ERA5 baseline (see Figure \ref{tables}, middle table).

Interestingly, while total CRPS values are lowest in the extratropics, relative improvements are largest in the tropics — particularly over the Atlantic and Pacific oceans (Figure~\ref{nts_maps}, middle column). Improvements are highest along the diagonal (i.e., when training and evaluation regions match), but extending the training region can also mitigate performance drops. For example, including the Sahara in the training domain improves performance in that region, demonstrating the value of broader training coverage.

\paragraph{Performance Relative to Direct Training on the Target}

The right column of Figure~\ref{tables} and the rightmost plots in Figure~\ref{nts_maps} compare model performance to directly training on the target region. As expected, models trained directly on the target achieve the best scores. However, some consistent transferability patterns emerge: models trained in extratropical regions generalize reasonably well to the tropics, whereas the reverse transfer (tropics $\rightarrow$ extratropics) results in larger performance drops.

Interestingly, northern-hemisphere models transfer better to the tropics than southern-hemisphere ones, and vice versa. This is despite the southern and northern extratropics being climatologically more similar to each other than either is to the tropics. Enlarging the training domain to include both the target and surrounding regions also improves performance — with up to a 5\% gain observed for the southern region.

\begin{figure*}[htb]
\begin{center}
\includegraphics[width=\textwidth]{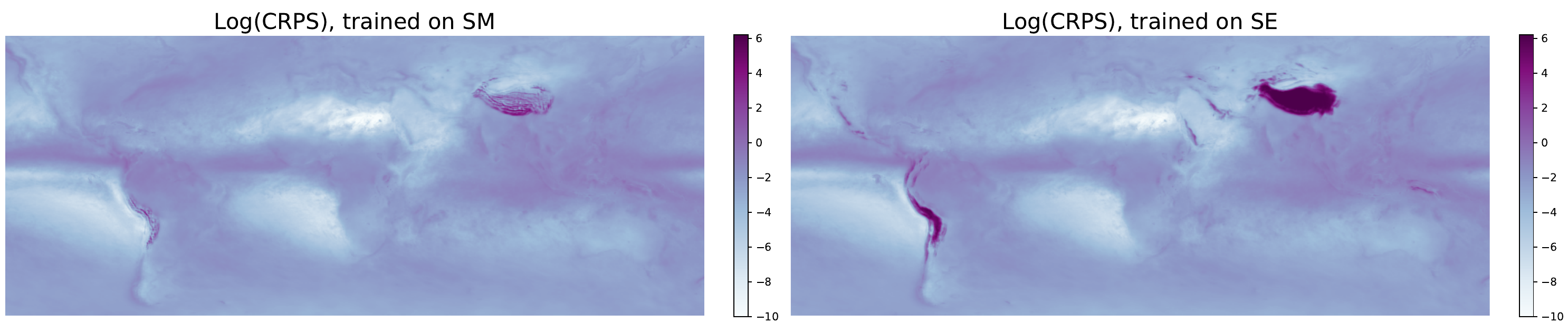}
\caption{The logarithm of the CRPS score is shown globally, averaged over the whole test period. On the left it shows the performance of the GAN trained on SM on the right trained on SE.}
\label{crps_log_maps}
\end{center}
\end{figure*}

%table
\begin{table*}[htb] 
\caption{Ablation study comparing models trained with and without high-resolution geographic features. CRPS (lower is better) for precipitation [mm/h] is shown for each evaluation region. Values represent the mean pixel-wise CRPS across 8 ensemble members. Best scores are in bold; failures are highlighted in red.}
\label{ablation_orography}
\vskip 0.15in
\begin{center}
\begin{small}
\begin{sc}
\begin{tabular}{l|c|c|ccccccccc}
\toprule
 %&  & & & & Target & Areas \\
% \multicolumn{4}\\
Train & train& inf. & \multicolumn{9}{c}{Eval Areas}\\
 Area &  geo & geo & NW & NM & NE & TW & TM & TE & SW & SM & SE \\
  \midrule
SM & $\checkmark$ & $\checkmark$ & 0.091 & \bf{0.052} & \textcolor{red}{0.549} & 0.146 & \bf{0.083} & \bf{0.220} & 0.090 &  0.075 & 0.074\\ 
%SM & $\checkmark$ & $\times$ & 0.097 & 0.062 & 0.147 & \textcolor{red}{0.300} &  0.090 & 0.263 & 0.089 & 0.083 & 0.081 \\
SM & $\times$ & $\times$ & \bf{0.090} &  0.056 & \bf{0.126} & \bf{0.141} & 0.085 & 0.247 & \bf{0.081} & \bf{0.074} & \bf{0.072}\\ 
\midrule
SE & $\checkmark$ & $\checkmark$ & 0.094 &  0.066 & \textcolor{red}{26.0} & \textcolor{red}{1.40} & 0.087 &  \bf{0.227} &  \textcolor{red}{0.636} &  \bf{0.074} &
        \bf{0.071}\\ 
%SE & $\checkmark$ & $\times$ & 0.170  & \textcolor{red}{0.516} & \textcolor{red}{4.90} & \textcolor{red}{2.81} & 0.106 & 0.394 & \bf{0.084} & 0.080 & 0.075 \\ 
SE & $\times$ & $\times$ & \bf{0.093} & \bf{0.053} & \bf{0.136} & \bf{0.175} & \bf{0.084} & 0.232 & \bf{0.085} & 0.075 & 0.072\\ 
\midrule
NW & $\checkmark$ & $\checkmark$ & \bf{0.078} & \bf{0.046} & \bf{0.092} & \bf{0.120} & \bf{0.082} & 0.223 &  \bf{0.081} & \bf{0.078} & \bf{0.076}\\
NW & $\times$ & $\times$ & 0.081&  0.047 & \bf{0.092} & 0.124 & 0.084 & \bf{0.217} & 0.085 & 0.079 & 0.080\\
\bottomrule
\end{tabular}
\end{sc}
\end{small}
\end{center}
\vskip -0.1in
\end{table*}

\paragraph{Performance Failures: SM and SE}

Severe performance degradation occurs when transferring models trained in SM or SE to high-elevation regions (e.g., NE, TW, TE, and SW). Both SM and SE are characterized by low elevation, and their models fail when applied to regions dominated by complex topography. These findings indicate that terrain-related processes, which are not well represented in the training data, pose a significant out-of-distribution challenge. In Figure \ref{crps_log_maps} we see that the very high CRPS values are located in areas like the Himalayas and Andes.

\paragraph{The Role of Orography}

To better understand the role of static geographic features — here orography and land–sea mask — we conduct ablation experiments in which these inputs are excluded during training and inference (Table~\ref{ablation_orography}).

For models trained in low-elevation regions (SM and SE), removing high-resolution geographic features alleviates several model failures (marked in red). This suggests that when the training domain lacks topographic variability it can harm generalization to mountainous regions. For other, less mountainous target areas, including high-res features is not improving performance when training on SE and SM. The only exception is TE, where including the land–sea mask slightly improves performance, likely due to the region’s many islands.

Conversely, for models trained in high-relief regions such as NW, including high-resolution geographic inputs generally improves performance across most evaluation domains. Here, interestingly, TE is the only target area where there are better results without high-res input features.

\begin{figure*}[htb]
\centering
\includegraphics[width=\textwidth]{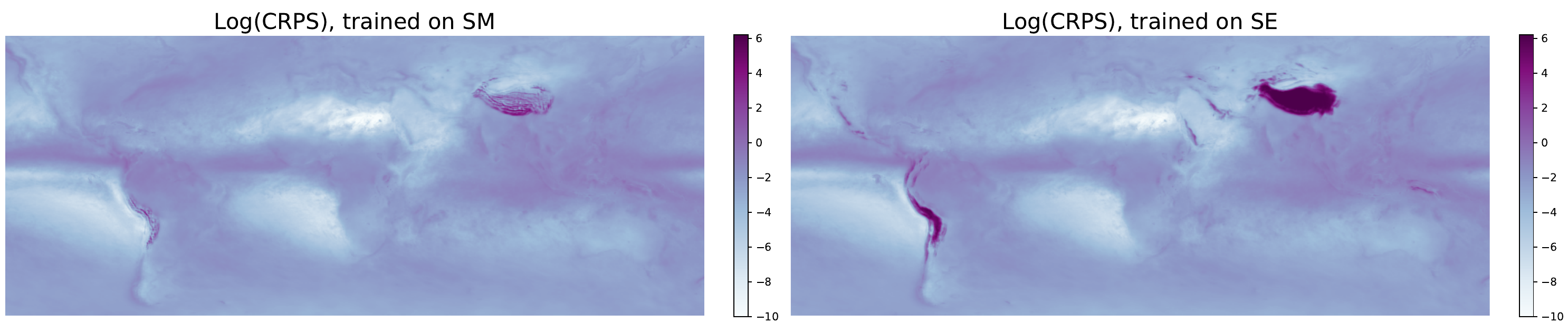}
\caption{Global CRPS (log scale) averaged over the test period for models trained on SM (left) and SE (right). Strong performance degradation occurs in mountainous regions such as NE and TW.}
\label{crps_log_maps}
\end{figure*}

\section{Conclusions}

This work demonstrates the successful application of a WGAN-based model to a novel precipitation downscaling task using IMERG as the target. Our results show strong global performance and clear improvements over ERA5 when transferring the model to new locations: in 12 out of 16 cases, the WGAN outperforms the baseline, achieving performance gains of $23–43\%$ in 11 of them. This suggests that machine learning downscaling models can, under certain conditions, be applied beyond their training domains without additional adaptation.

However, performance degradation remains a significant challenge when transferring to unseen regions. In four cases, the model fails entirely, and in others, we observe performance drops of up to $46\%$, with typical decreases ranging from $3 - 25\%$. Orography in particular introduces substantial out-of-distribution challenges, indicating that it may be better to exclude regions with largely different topography when deploying models trained elsewhere or dropping topography as a feature. Beyond these low-to-high elevation transitions, the tropics present the most difficult area to transfer from. Models trained on extratropical regions tend to generalize better to tropical climates than the reverse, and those trained in the Northern Hemisphere transfer more effectively to the tropics than those trained in the south.

Further analysis could investigate the temporal distribution of the CRPS to assess whether model performance is dominated by individual events, as suggested by some sharp spatial patterns. Complementary evaluation metrics, especially those targeting extremes, would also provide a more comprehensive understanding of model behavior. Future work could extend this study by exploring alternative architectures, such as diffusion-based models, and by moving beyond ERA5 reanalysis data to IFS hindcasts. Incorporating location-restricted radar observations in place of globally available satellite data could further enhance the relevance and applicability of downscaling approaches.

%\section*{Acknowledgments}
%This was was supported in part by......

%Bibliography
%\bibliographystyle{unsrt}  
\bibliographystyle{plainnat}
\bibliography{references}

\end{document}